\documentclass{article}

\usepackage{PRIMEarxiv}

\usepackage[utf8]{inputenc} 
\usepackage[T1]{fontenc}    
\usepackage{tabularx}
\usepackage{comment}
\usepackage{multirow}
\usepackage{amsmath}
\usepackage{graphicx}
\usepackage{natbib}
\usepackage{hyperref}
\usepackage{subfig}
\usepackage{siunitx}
\usepackage{xcolor}
\usepackage[normalem]{ulem}
\usepackage{siunitx}
\usepackage{listings}
\usepackage{latexsym}
\usepackage{amsfonts}
\usepackage{amsmath}
\usepackage{graphicx}
\usepackage{qtree}
\usepackage{pict2e}
\usepackage{eepic}
\usepackage{color}
\usepackage{moreverb}
\usepackage{fancyvrb}
\usepackage{tabularx}
\usepackage{colortbl}
\usepackage{acronym}
\usepackage{multirow}
\usepackage{longtable}
\usepackage{paralist}
\usepackage{url}
\usepackage{xspace}
\usepackage{multirow}			
\usepackage{algorithm}
\usepackage{algorithmicx}
\usepackage{algpseudocode}
\usepackage{verbatim}			
\usepackage{array}				
\usepackage{colortbl}			
\usepackage{tikz-qtree}
\usepackage{rotating}
\usepackage{indentfirst}
\usepackage{booktabs}
\usepackage{epstopdf}
\usepackage{bm}
\usepackage{natbib}

\title{Towards a fully Unsupervised framework for intent induction in Customer Support Dialogues}

\author{Rita Costa\\
INOV/Instituto Superior Técnico,\\
\texttt{ritaflscosta@tecnico.ulisboa.pt}
\And
Bruno Martins\\
INESC-ID/Instituto Superior Técnico,\\
\texttt{bruno.g.martins@tecnico.ulisboa.pt}
\And
Sérgio Viana\\
Xpand-it,\\
\texttt{sergio.viana@xpand-it.com}
\And
Luisa Coheur\\
INESC-ID/Instituto Superior Técnico,\\
\texttt{lcoheur@edu.ulisboa.pt}}

\begin{document}
\maketitle

\begin{abstract}
State of the art models in intent induction require annotated datasets. However, annotating dialogues is time-consuming, laborious and expensive. In this work, we propose a completely unsupervised framework for intent induction within a dialogue. In addition, we show how pre-processing the dialogue corpora can improve results. Finally, we show how to extract the dialogue flows of intentions by investigating the most common sequences. Although we test our work in the MultiWOZ dataset, the fact that this framework requires no prior knowledge make it applicable to any possible use case, making it very relevant to real world customer support applications across industry.
\end{abstract}

\section{Introduction}
\label{sec:intro}

The evolution of technology has allowed the automation of several processes across diversified engineering industry fields, such as customer support services, which have drastically evolved with the advances in Natural Language Processing and Machine Learning. One of the major challenges of these systems is to identify users intentions, a complex Natural Language Understanding task, that vary across domains. With the  evolution of Deep Learning architectures, recent works focused on modelling intentions and creating a taxonomy of intents, so they can be fed to powerful supervised clustering algorithms \citep{Haponchyk2020, Chatterjee2021}. 

However, these systems have the bottleneck of requiring the existence of labelled data to be trained and deployed, and, thus, they can not be easily transferred to real world customer support services, where the available data for a commercial chatbot usually consists in no more than a dataset of interactions between clients and operators. As labeling hundreds of utterances with intent labels can be time-consuming, laborious, expensive and, sometimes, even requires someone with expertise, it is not straightforward to apply current state of the art supervised models to new domains \citep{Chatterjee2020}.

In this work, we propose a totally unsupervised intent induction framework and apply it to the MultiWOZ dataset \citep{Budzianowski2019}. Previous unsupervised intent induction works have used methods which perform clustering of user query utterances in human-human conversations \citep{Perkins2019, Haponchyk2020, Chatterjee2020}. Popular clustering algorithms for practical applications include centroid-based algorithms, such as the K-Means algorithm \citep{Lloyd1982}, and density based algorithms, namely DBSCAN \citep{Daszykowski2009} and HDBSCAN \citep{McInnes2017}. An advantage of the density-based algorithms is not requiring to define the number of clusters a priori \citep{Ghaemi2019}, being more efficient for detecting clusters with arbitrary shapes from a noisy dataset, particularly for a case where the number of dialogue intentions is not known a priori. By using HDBSCAN, we also do not require the prior definition of the density threshold used to create the clusters (contrary to DBSCAN), which is more suitable for this application. Moreover, we show that text pre-processing techniques, such as performing named entity recognition, can improve the clustering process of dialogue utterances. Finally, we complement this experiment with an analysis of the most common dialogue flows, based on the detected intents. 

In summary, the main contributions of this work are: 

\begin{itemize}
\item the application of a fully unsupervised method for extracting intents within a dialogue, requiring no prior information about its content, and hence avoiding the time-consuming task of manually analysing user questions and {identifying the intents} (both intra- and inter-domain studies are conducted);

\item an exploratory analysis of the dataset, motivating the usage of {general text processing techniques to optimize the intent extraction process, that can be applied to any corpora;} 
    
\item an informal analysis of the most common flows of discovered intentions.
\end{itemize} 

As there is no required prior knowledge of any dataset specificities or deployment details, our proposal is applicable to any type of data and use case, making it relevant for a huge variety of applications, such as customer support applications. 

This paper is organized as follows: in Section \ref{sec:related_work} we present related work, in Section \ref{section:preliminary_data_analysis} we present a data analysis, and in Section \ref{sec:experimental_results} the experimental results. Then, in Section \ref{section:ConclusionsAndFutureWork} we present the main conclusions and some future work.

\section{Related Work}
\label{sec:related_work}

This section gives an overview of the tools used in the development of this work. In Section \ref{subsec:multiwoz}, we present the MultiWOZ dataset, a task-oriented collection of dialogues whose utterances are used for the experiments in Section \ref{sec:experimental_results}. Before feeding these sentences into an algorithm, it is required to transform them in a space representation, for which an overview is given in Section \ref{subsection:text_representation}. In Section \ref{subsec:hdbscan}, we present HDBSCAN and motivate the choice of this clustering algorithm. Finally, the method for analysis of dialogue flows is presented in Section \ref{subsec:sequential_pattern_mining}.

\subsection{MultiWOZ Dataset}
\label{subsec:multiwoz}

The MultiWOZ dataset \citep{Budzianowski2019} is a labelled human-human collection of goal-oriented dialogues, simulating natural conversations between a tourist and an assistant from an information center in a touristic city. The corpus has conversations spanning over 7 domains --- \textit{attraction, hospital, police, hotel, restaurant, taxi, train} --- with diverse complexity of tasks, going from a simple information query about an attraction, to booking a night at a hotel, a restaurant reservation and a taxi to connect both places. The dataset is composed of 10438 dialogues, which can be either single domain or multi-domain. The average number of turns per dialogue is 8.93 and 15.39, for single and multi-domain, respectively. 

\begin{figure}[!htb]
  \centering
  \includegraphics[width=0.7\textwidth]{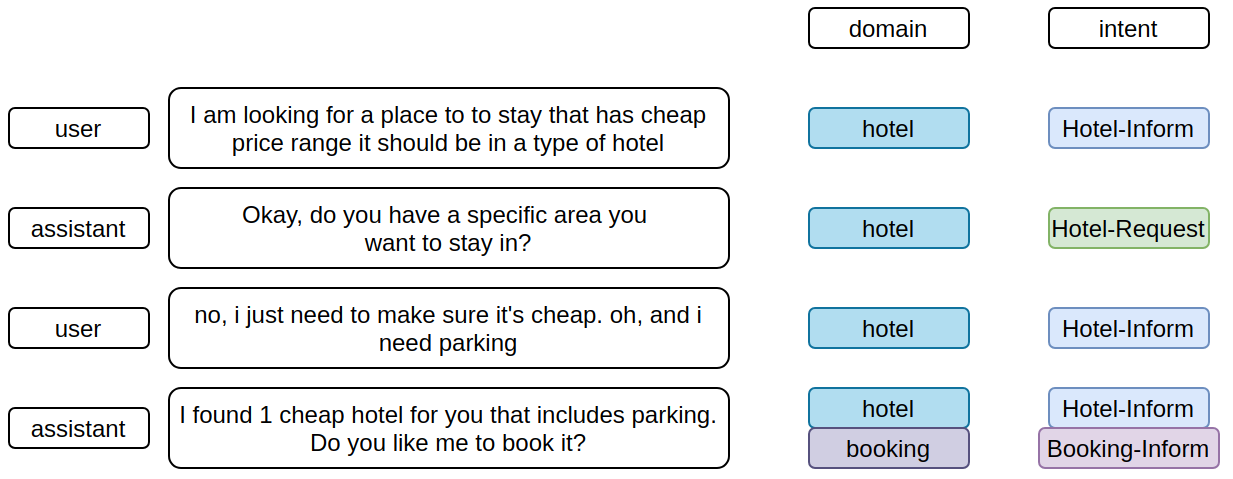}
  \caption{A dialogue example with domains and intents.}
  \label{fig:dialogue_example}
\end{figure}

One particularity about this dataset is the richness in annotations at two levels for each utterance: domain and intent. This information will allows us to conduct the experiments with a reference of ground truth, helping validating the used approach. In Figure \ref{fig:dialogue_example}, it is possible to see an example of a part of a dialogue, with the corresponding domains and intents for each utterance. Besides the possible conversation domains, an utterance can also belong to two broader domains: the \textit{booking} domain --- if it refers to the act of booking an entitiy --- or to the \textit{general} domain --- if it is a greeting, an acknowledgement, etc. In addition to the dialogues and their annotations, the dataset is also composed of 7 database files, one for each possible domain of conversation. 

A further exploratory analysis of this dataset can be found in Section \ref{section:preliminary_data_analysis}.

\subsection{Text Representation}
\label{subsection:text_representation}

An important part of Natural Language Processing is how to represent sentences such that it is possible to build algorithms on them. Initially, the focus was in representing words independently. The most basic approach was to represent text through a one-hot vector, with value 1 assigned to a word that is present, and 0 corresponding to not present. The impossibility to transmit the similarity between words gave rise to what are now called word embeddings, which represent a word in a low dimensional vector space, where similar words take similar parts of the modelling space. Popular word embeddings techniques include Word2Vec  \citep{Mikolov2013} and GloVe \citep{Pennington2014}. The need to solve ambiguities around words meanings and represent them with respect to the sentence they are inserted led to the evolution of contextual word embeddings, such as ELMo \citep{Peters2018}. The representations move beyond word-level semantics, in that each word has a representation which is a function of the entire input sequence, being able to capture syntax and semantics. The evolution of text representation techniques opened the door to more complex language models, with transformer architectures that use attention to learn embeddings, such as GPT \citep{Radford2018} and BERT \cite{Devlin2018}. In tasks such as clustering and semantic search, a common method is to map each sentence such that semantically similar sentences are close, as proposed in Sections \ref{subsubsec:sentence-bert} and \ref{subsubsec:dimensionality_reduction}.

\subsubsection{Sentence-BERT}
\label{subsubsec:sentence-bert}

BERT related models have the state-of-the-art performance on sentence-pair regression tasks like semantic textual similarity. However, to compute the similarity between two sentences requires that they are both fed into the model, which makes it too expensive for pair regression tasks, such as semantic similarity search and clustering, due to too many possible combinations. To make this task more efficient, \textbf{Sentence-BERT} (SBERT) \citep{Reimers2020} uses siamese and triplet network structures to derive semantically meaningful sentence embeddings. These techniques represent entire sentences and their semantic information as vectors, making semantically similar sentences close in the vector space. This helps the machine in understanding the context, intention, and other nuances in the entire text. Then, by using a similarity measure like cosine-similarity or euclidean distance, it is possible to find semantically similar sentences. SBERT is available in the Sentence-Transformer framework\footnote{\url{https://www.sbert.net/}}, with pre-trained models of sentence embeddings tuned for various tasks, in more than 100 languages.

\subsubsection{Dimensionality Reduction}
\label{subsubsec:dimensionality_reduction}

After using SBERT for utterance representation, we obtain embeddings with dimension of 768. Since high dimensionality embeddings lead to a loss of robustness in clustering algorithms, we trade the loss of information for a more robust clustering by reducing the dimensionality of the embeddings before feeding them to the clustering algorithm.

There are a few alternatives of methods for dimensionality reduction, such as t-Distributed Stochastic Neighbor Embedding (\textbf{t-SNE}) \citep{Maaten2008}, and Uniform Manifold
Approximation and Projection (\textbf{UMAP}) \citep{McInnes2018}. Both were designed to predominantly preserve the local structure, by grouping neighbouring data points together, which provides a very informative visualization of the heterogeneity present in the data. UMAP is more adequate for this context, since t-SNE produces unstable embeddings, making the experiences non reproducible.

\subsection{HDBSCAN for unsupervised clustering}
\label{subsec:hdbscan}

Clustering is an unsupervised Machine Learning technique that consists of grouping data points such that those with similar features are classified with the same group, meaning that data points belonging to different groups should have more dissimilar properties. Depending on the notion of what defines a cluster, there are a variety of diversified clustering algorithms: some of them are centroid based, such as the K-Means \citep{Lloyd1982}, where the clustering is done based on some randomly initialized points and the minimum distance from a point to others; others are density based, such as DBSCAN \citep{Daszykowski2009}, where points are clustered based on their densities in a particular region. Density based clustering is particularly relevant for problems where little is known about the dataset, since they do not require the a priori definition of the amount of clusters.

In most density-based clustering algorithms, such as DBSCAN, it is necessary to define a density threshold to make a cluster. This parameter is specially difficult to adjust for higher dimensional data, posing a problem for obtaining clusters with varying densities. To solve this problem, the Hierarchical Density-Based Spatial Clustering of Applications with Noise (\textbf{HDBSCAN}) \citep{Campello2015} was developed, not requiring the prior definition of this density threshold. The algorithm first builds a hierarchy to figure out which peaks end up merging together and in what order. Then, for each cluster, it evaluates if it is more beneficial to keep that cluster or split it into subclusters, considering the volume of each peak. HDBSCAN uses soft clustering: unlike most clustering algorithms, data points are not assigned cluster labels, but rather a vector of probabilities for each cluster, identified by $c \in \{0, n_{clusters}-1\}$, allowing each point to potentially be a mix of clusters. It is also noise aware, meaning that it has a notion of data samples that are not assigned to any cluster, to which it assigns the label -1. 

\subsection{Sequential Pattern Mining for the analysis of dialogue flows}
\label{subsec:sequential_pattern_mining}

In the context of dialogue interactions, besides identifying utterances intentions, it is relevant to evaluate the most common interactions, allowing to discover the flow of the dialogue. To do so, it is possible to use the sequential pattern mining algorithm \textbf{Prefix}-projected \textbf{S}equential \textbf{pa}tter\textbf{n} (\textbf{PrefixSpan}) \citep{Pei2001}, which discovers frequent subsequences as patterns in a sequence database.

The PrefixSpan implementation\footnote{\url{https://pypi.org/project/prefixspan/}} to be used outputs traditional single-item sequential patterns. This library also includes the frequent closed sequential pattern mining algorithm BIDE \citep{Wang2004}, and the frequent generator sequential pattern mining algorithm FEAT \citep{Gao2008}. To use the algorithm via API, we will refer to the PrefixSpan class in \url{prefixspan/api.py}. In this implementations, two types of sequences can be obtained: \verb|.frequent(n)| returns the sequences that appear n times; and \verb|.topk(k)| gives the k most frequent sequences in the dataset. These methods also support other specificities, which can be found in the algorithm documentation link. 
    
\section{Data Analysis}
\label{section:preliminary_data_analysis}

To have a better understanding of the task we have at hand, it is relevant to perform an analysis of the dialogue utterances. In Section \ref{subsec:pda_embeddings}, the similarities between embeddings of different utterances are investigated, motivating the use of an open-source tool for entity identification. In Section \ref{subsec:pda_domain_intent}, we provide an overview of the distribution of the dataset over domain and intent. 

\subsection{Embeddings representation}
\label{subsec:pda_embeddings}

As proposed in Section \ref{subsection:text_representation}, the dataset utterances are represented using the a Sentence-BERT model. These embeddings are obtained using the sentence-transformer \verb|`paraphrase-distilroberta-base-v1'| package, that outputs embeddings with dimension 768 for each utterance. 

\begin{figure}[h!]
	\centering
	\subfloat[Similarity between pairs of embeddings. \label{fig:pairs_similarity}]{\includegraphics[width=0.49\linewidth]{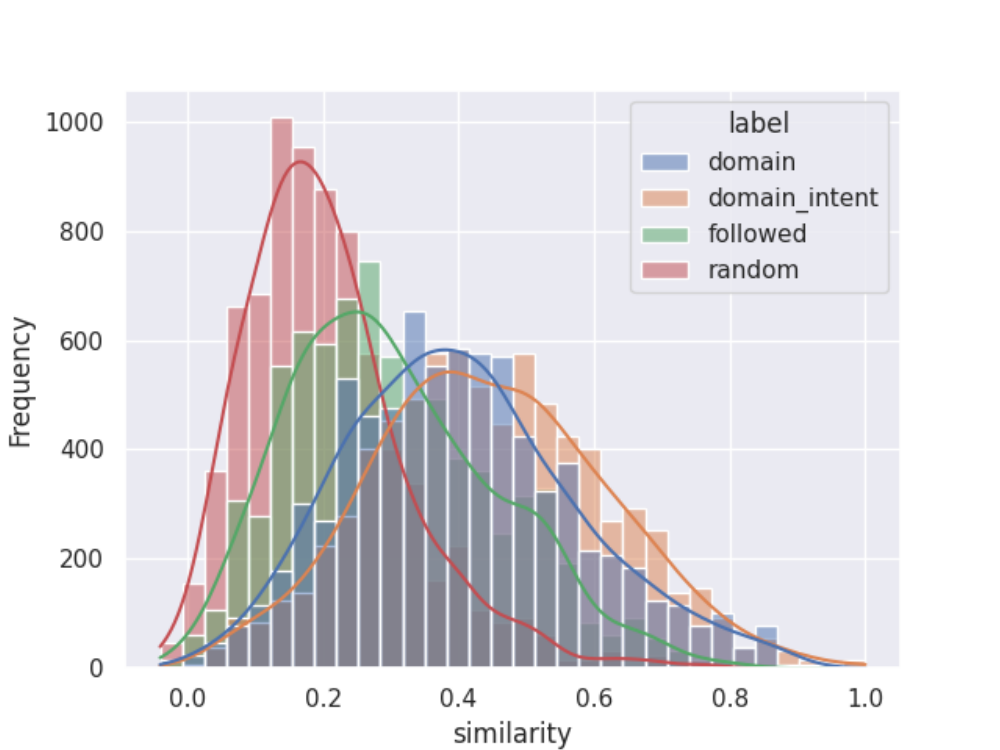}}%
	\subfloat[Similarity between pairs of embeddings after NER tool from spacy. \label{fig:pairs_similarity_spacy_ner_original}]{\includegraphics[width=0.49\linewidth]{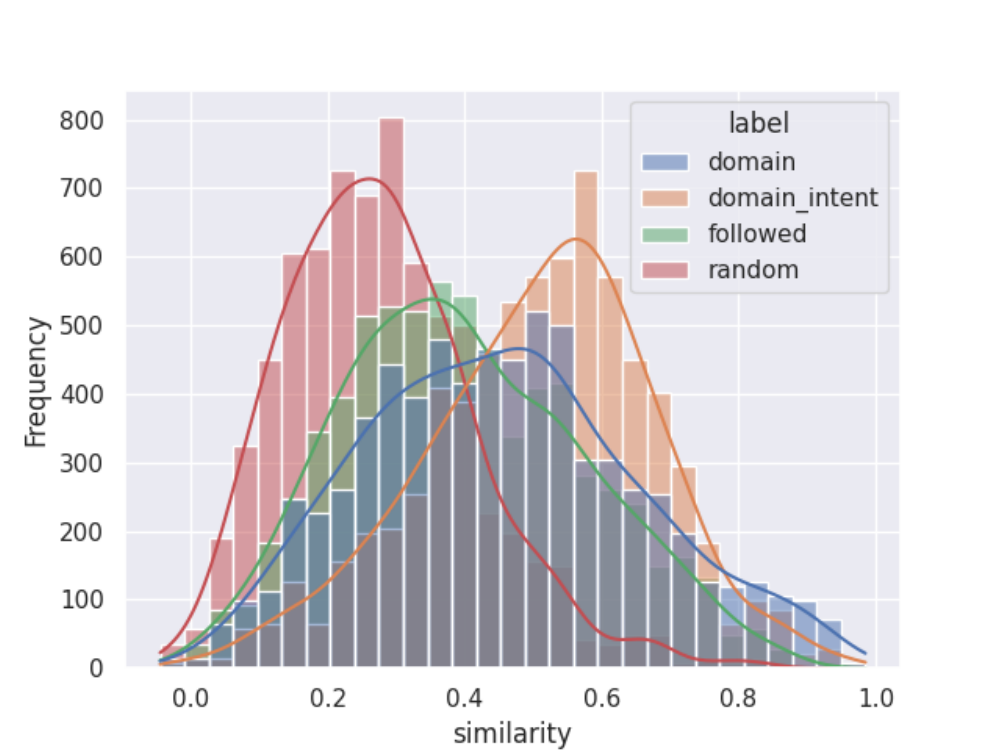}}%
	
	\subfloat[Similarity between pairs of embeddings after adding entities to the NER tool from spacy. \label{fig:pairs_similarity_added_entities}]{\includegraphics[width=0.49\linewidth]{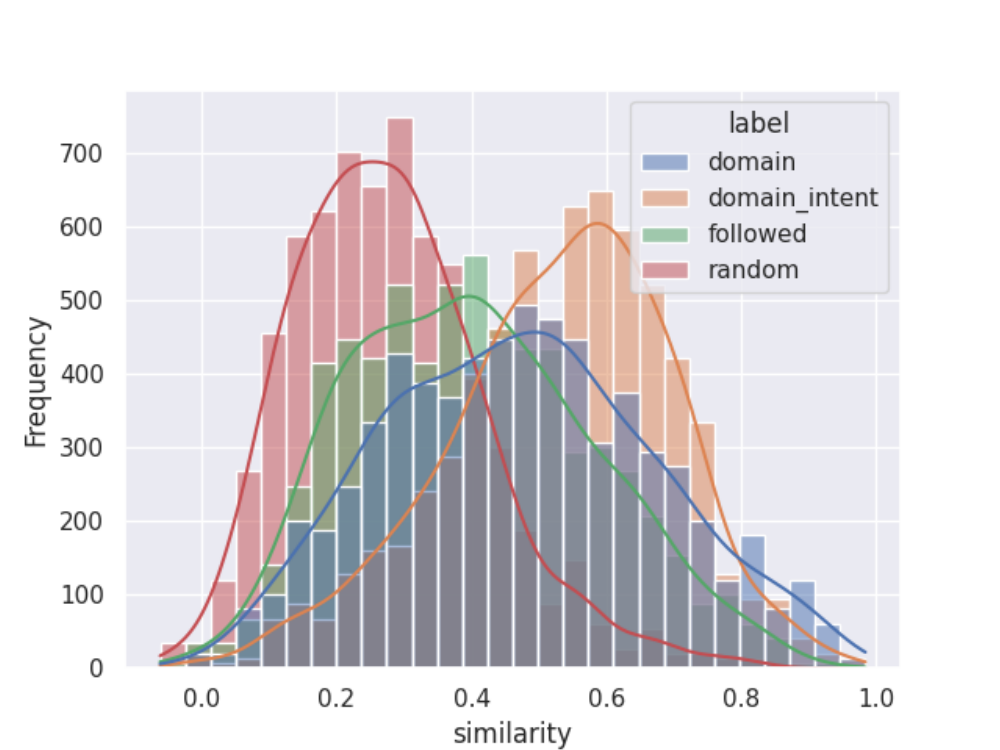}}%
	\caption{Similarity.}
	\label{fig:pairs_similarityGeneral}
\end{figure}

Following \cite{Laban2021}, to evaluate the feasibility of this experience, we measured the similarity between the embeddings of 1000 pairs of random utterances belonging to the different categories: 
\begin{itemize}
    \item utterances belonging to the same domain, or group of domains (label \verb|domain|); 
    \item utterances belonging to the same domain and the same intent, or groups of domains and intents (label \verb|domain_intent|); 
    \item subsequent utterances in the dialogue (label \verb|followed|); 
    \item and utterances randomly obtained from the dataset (label \verb|random|). 
\end{itemize} 

The plot can be seen in Figure \ref{fig:pairs_similarity}. As the goal is to discover broader intents for the dialogues utterances, it can be useful to make them more homogeneous, in order to avoid clusters around entities. For this purpose, we use the spaCy Named Entity Recognition tool\footnote{\url{https://spacy.io/usage/linguistic-features}}, which replaces the recognized entities for broader tags, such as {general known numbers (e.g one, thirty-three, two hundred) and places (e.g Cambridge)}. The similarity plot for these processed utterances is shown in Figure \ref{fig:pairs_similarity_spacy_ner_original}. In addition to this, as there is information about specific entities present in the dataset, such as hotel and restaurants names, we can also remove them from the dialogues (e.g. In Figure \ref{fig:pairs_similarity_added_entities}, we can see the similarity between pairs of utterances with entities from the dataset removed.

From the plots, it is clear that the distribution of similarity between pairs of embeddings is higher for utterances with common domains and intentions, suggesting that a clustering experience based on this measure may be successful. Besides, this difference is higher for utterances where entities were removed, motivating the use of this tool for improving the clustering experience.

\subsection{Domain and intent annotations}
\label{subsec:pda_domain_intent}

\begin{figure}[!htb]
  \centering
  \includegraphics[width=1\textwidth]{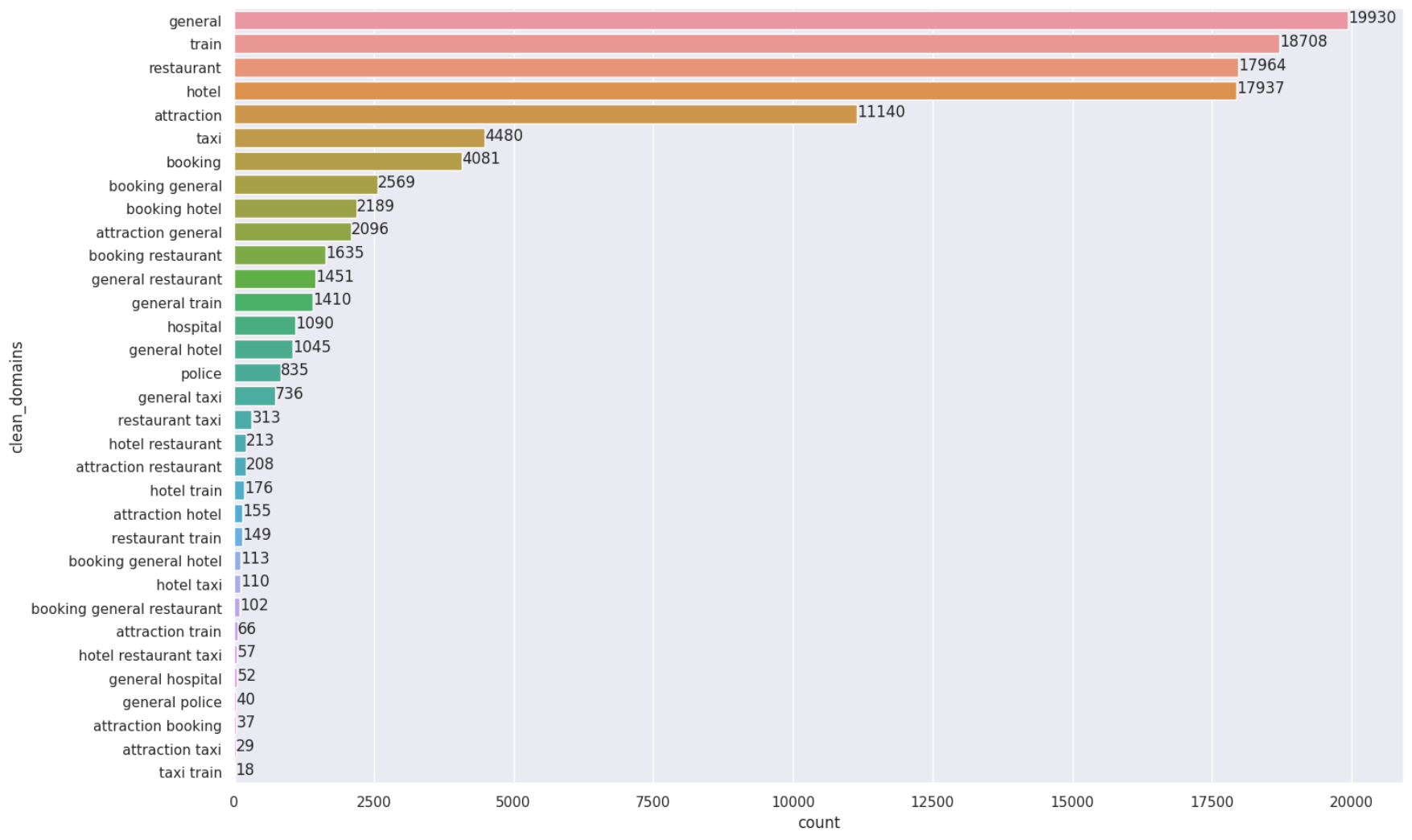}
  \caption{The possible combinations of domains.}
  \label{fig:all_domains_combinations}
\end{figure}

As seen in Figure \ref{fig:dialogue_example}, one utterance from the MultiWOZ dataset can belong to more than one domain. In Figure \ref{fig:all_domains_combinations}, we present the frequency of each possible combination of domains (the combinations which were present less than 10 times were kept out of this plot for the sake of visibility). The highest presence are for single domain utterances. The highest value is for the \textit{general} domain, followed by \textit{train}, \textit{restaurant}, \textit{hotel}, \textit{attraction}, \textit{taxi} and \textit{booking}. The \textit{police} and \textit{hospital} domains have fewer utterances assigned, as they also have less dialogues. 

\begin{figure}[!htb]
  \centering
  \includegraphics[width=0.8\textwidth]{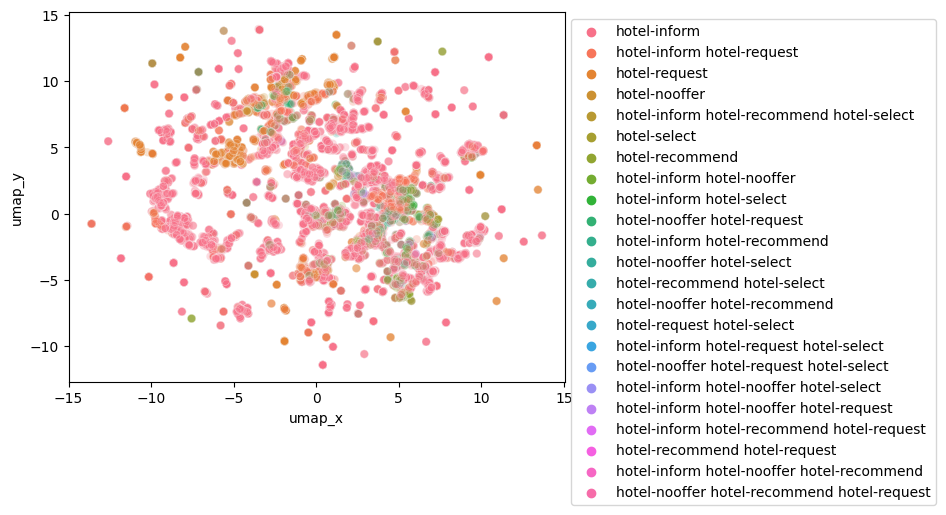}
  \caption{2-D representations of utterances embeddings per intent in the hotel domain.}
  \label{fig:hotel_intents_embeddings}
\end{figure}

For the generality of domains, the possible intents classifications are \textit{inform}, \textit{request}, \textit{recommend}, \textit{no-offer} and \textit{select}. The \textit{booking} domain has other possibilities regarding the booking possibility, \textit{book} or \textit{no-book}. The \textit{general} domain has its particular intents: \textit{greet}, \textit{welcome}, \textit{reqmore}, \textit{thank} and \textit{bye}. Naturally, it is possible for a utterance to hold more than one intent. As there are many more possible combinations of intents than domains, we will not present plots of all domains, but rather exemplify with the representations of utterances belonging to the hotel domain. In Figure \ref{fig:hotel_intents_embeddings}, it is possible to see the 2-D representations of utterances belonging to this domain, using the UMAP algorithm with dimension 2. Although these are just 2-D representations of way higher dimensional embeddings, it is still possible to identify some groups of sentences belonging to the same domain or intent. This can translate the possibility of performing density clustering in these data points. 

\section{Experimental Results}
\label{sec:experimental_results}

This section includes an analysis of the experimental results. An introduction to the evaluation methods is given in Section \ref{subsec:evaluation_metrics}. In Section \ref{subsec:intradomain_clustering}, we present and analyse the results of an intra-domain clustering experiment for the hotel domain. In Section \ref{subsec:interdomain_clustering}, an inter-domain clustering experience is conducted.


\subsection{Evaluation Metrics}
\label{subsec:evaluation_metrics}

To evaluate the results of a clustering experiment, one can use intrinsic methods (based on properties of the algorithm itself), such as the relative validity index. This metric measures how close elements from one cluster are to each other, and how distant they are from elements in other clusters. It is important to note that the topic of clustering validation is considered one of the most challenging topics in the clustering literature: since these are unsupervised algorithms, it is required to resort to internal validity criteria, calculated solely based on information intrinsic to the data.

In these particular experiences, since we have annotation references from the dataset, it is also possible to resort to extrinsic methods that compare the clusters with a pre-existing structure --- a ground truth solution. In this context, BCubed precision and BCubed recall \citep{Bagga1998} are found to be the only ones that satisfy all the proposed properties/constraints for clustering evaluation metrics \citep{Amigo2009}. The BCubed precision of an item is the proportion of items in its cluster which have the item’s category, including itself, related to the amount of items in its cluster. The BCubed recall is analogous, but related to the amount of items within its category. The overall BCubed precision and recall are the averaged precision and recall of all the items. 

Naturally, extrinsic methods are not usable when there are no ground truth references, leaving intrinsic methods as the most relevant for clustering experiences, since they are the only ones applicable in real world scenarios.

\subsection{Inter-domain Clustering}
\label{subsec:interdomain_clustering}

\begin{figure}[!htb]
  \centering
  \includegraphics[width=\textwidth]{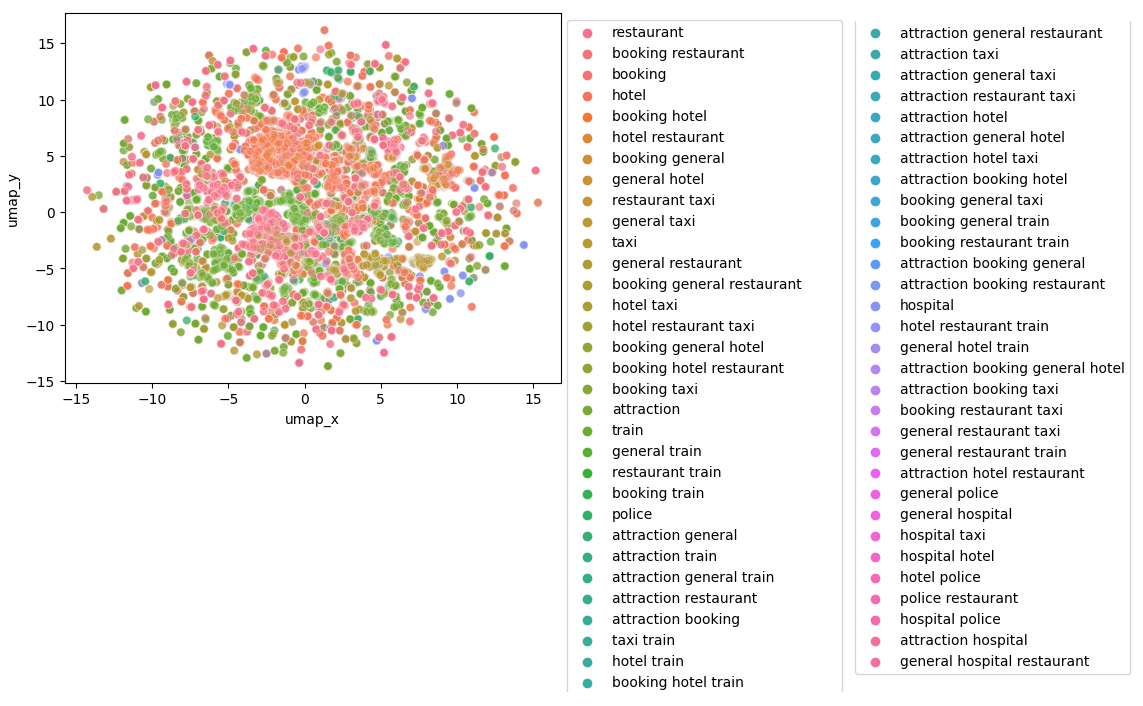}
  \caption{2-D representations of utterances embeddings per domain.}
  \label{fig:multiwoz_embeddings}
\end{figure}

Firstly, we present the clustering results for an experience with all the utterances from the MultiWOZ dataset. In this inter-domain clustering experience, we have two types of possible labels: domain and intent. To simplify, we present the possible domains for the utterances, whose 2-D representations are plotted in Figure \ref{fig:multiwoz_embeddings}. As evident in Figure \ref{fig:multiwoz_embeddings}, there are many possible combinations of domain labels for the data points. Hence, we will refrain from plotting the possible combinations of intents, as there are even more possibilities than those in Figure \ref{fig:multiwoz_embeddings}, and its analysis would be too exhaustive. 

For these experiences, we opted to remove the utterances from the general domain. As seen in the plot from Figure \ref{fig:all_domains_combinations}, these are the most present in the dataset. The fact that these types of utterances are very repetitive, with a very low variability, makes the dataset very imbalanced. As the same exact utterance from the general domain occurs very often, this can damage the clustering experience by generating clusters for equal utterances only, which is not aligned with the goals for this task.

When running the HDBSCAN algorithm, there are two important parameters to set: \verb|min_cluster_size|, defining the smallest size grouping that should be considered a cluster. The bigger its value, the less clusters will be obtained; and \verb|min_samples|, which provides a measure of how conservative the clustering should be. The larger its value, the more conservative the clustering, and the more points will be considered as noise, being clusters progressively restricted to more dense areas. By default, it is set to the value of \verb|min_cluster_size|. We fine tune these values by making \verb|min_samples| vary from 0 to 100 with a step size of 10, \verb|min_cluster_size| vary from 25 to 300 with a step size of 25 and measuring the relative validity index, as depicted in Table \ref{tab:finetune_alldomains_mwoz_validity}. It is not possible to see a direct relationship between this value and both of the variables. The best result happens for $\verb|min_samples|=100$ and $\verb|min_cluster_size|=300$.

\begin{table}[!htb]
  \renewcommand{\arraystretch}{1.2} 
  \centering
  \caption{Grid search {of relative validity index} over min\_cluster\_size and min\_samples for all utterances of the MultiWOZ dataset.}
  \resizebox*{\textwidth}{!}{%
  \begin{tabular}{lcccccccccccc}
    \hline
     & 25 & 50 & 75 & 100 & 125 & 150 & 175 & 200 & 225 & 250 & 275 & \textbf{300} \\
    \hline
        10 & \num{3.32e-2} & \num{4.31e-2} & \num{3.70e-5} & \num{2.47e-5} & \num{2.47e-5} & \num{2.47e-5} & \num{3.09e-6} & \num{3.09e-6} & \num{3.09e-6} & \num{3.09e-6} & \num{3.09e-6} & \num{3.09e-6}\\
        20 & \num{4.01e-2} & \num{3.76e-2} & \num{2.17e-3} & \num{1.50e-3} & \num{1.50e-3} & \num{4.95e-6} & \num{4.95e-6} & \num{2.02e-3} & \num{1.25e-4} & \num{1.23e-4} & \num{2.60e-3} & \num{2.60e-3}\\
        30 & \num{3.30e-2} & \num{2.67e-2} & \num{1.01e-4} & \num{1.38e-3} & \num{1.38e-3} & \num{1.38e-3} & \num{1.15e-5} & \num{5.46e-6} & \num{5.46e-6} & \num{4.38e-2} & \num{4.38e-2} & \num{4.15e-2}\\
        40 & \num{1.62e-2} & \num{1.62e-2} & \num{5.72e-5} & \num{8.88e-4} & \num{4.17e-6} & \num{1.26e-2} & \num{1.11e-2} & \num{9.46e-3} & \num{4.23e-2} & \num{4.03e-2} & \num{4.03e-2} & \num{4.03e-2}\\
        50 & \num{1.62e-2} & \num{2.48e-2} & \num{1.16e-3} & \num{5.89e-6} & \num{5.89e-6} & \num{6.99e-3} & \num{5.62e-3} & \num{1.44e-2} & \num{1.44e-2} & \num{5.10e-3} & \num{5.10e-3} & \num{5.09e-3}\\
        60 & \num{1.82e-2} & \num{2.06e-2} & \num{6.17e-4} & \num{2.42e-5} & \num{2.42e-5} & \num{8.76e-3} & \num{7.50e-3} & \num{3.88e-3} & \num{3.88e-3} & \num{3.88e-3} & \num{3.88e-3} & \num{3.87e-3}\\
        70 & \num{2.52e-5} & \num{2.52e-5} & \num{1.03e-3} & \num{1.03e-3} & \num{1.03e-3} & \num{2.02e-2} & \num{1.89e-2} & \num{1.66e-2} & \num{1.34e-2} & \num{1.14e-2} & \num{1.14e-2} & \num{3.09e-2}\\
        80 & \num{5.03e-4} & \num{5.03e-4} & \num{1.42e-5} & \num{1.42e-5} & \num{1.42e-5} & \num{1.87e-2} & \num{7.15e-2} & \num{7.15e-2} & \num{7.15e-2} & \num{7.15e-2} & \num{8.43e-2} & \num{1.04e-1}\\
        90 & \num{8.46e-7} & \num{5.81e-4} & \num{5.81e-4} & \num{1.14e-5} & \num{1.14e-5} & \num{8.93e-2} & \num{8.81e-2} & \num{8.57e-2} & \num{8.57e-2} & \num{9.97e-2} & \num{9.77e-2} & \num{9.80e-2}\\
        \textbf{100} & \num{3.04e-6} & \num{3.82e-6} & \num{3.82e-6} & \num{3.82e-6} & \num{3.82e-6} & \num{1.37e-1} & \num{1.36e-1} & \num{1.36e-1} & \num{1.12e-1} & \num{1.10e-1} & \num{1.10e-1} & \num{1.50e-1}\\
    \hline
  \end{tabular}
  }%
  \label{tab:finetune_alldomains_mwoz_validity}
\end{table}

\begin{table}[!htb]
    \renewcommand{\arraystretch}{1.2} 
    \centering
    \caption{Optimal clustering results for each value of min\_samples, with BCubed metrics computed using the domain labels.}
    \resizebox*{\textwidth}{!}{%
    \begin{tabular}{@{}lccccccccccccc@{}} 
    \hline
    & & & \phantom{abc} & \multicolumn{3}{c}{Clusters} & \phantom{abc} & \multicolumn{3}{c}{Soft Clusters} & \phantom{abc}& \\
    \cline{5-7}
    \cline{9-11}
    {min\_samples} & min\_cluster\_size & validity && P & R & M && P & R & M && n clusters\\
    \hline
    \textbf{10} & \textbf{50} & \num{4.31e-2} && 0.2559 & 0.6323 & \textbf{0.3643} && 0.5167 &	0.0915 & 0.1555 &&	\textbf{105}\\
    20 & 25 & \num{4.01e-2} && 0.2302 & 0.7171 & 0.3486 && 0.5236 & 0.0905 & 0.1544 && 159\\
    30 & 250 & \num{4.38e-2} && 0.2301 & 0.6409 & 0.3386 && 0.3884 & 0.3509 & 0.3687 && 16\\
    40 & 225 & \num{4.23e-2} && 0.2300 &	0.6583 & 0.3409 && 0.3877 & 0.3837 & 0.3857 && 15\\
    50 & 50 & \num{2.48e-2} && 0.2096 & 0.7328 & 0.3260 && 0.4733 & 0.1999 & 0.2811 && 55\\
    60 & 50 & \num{2.06e-2} && 0.2068 & 0.7232 & 0.3217 && 0.418 & 0.2506 & 0.3133 && 43\\
    70 & 300 & \num{3.09e-2} && 0.2066 & 0.6967 & 0.3187 && 0.3665 & 0.3625 & 0.3645 && 11\\
    80 & 300 & \num{1.04e-1} && 0.2202 & 0.6869 & 0.3335 && 0.3635 & 0.4073 & 0.3841 && 10\\
    \textbf{90} & \textbf{250} & \num{9.97e-2} && 0.2293 & 0.6770 & 0.3426 && 0.3833 & 0.3987 & \textbf{0.3909} && \textbf{12}\\
    \textbf{100} & \textbf{300} & \num{1.50e-1} && 0.2205 & 0.6689 & 0.3317 && 0.3473 & 0.4139 & 0.3777 && \textbf{9}\\
    \hline
    \end{tabular}
    }%
   \label{tab:finetune_alldomains_mwoz}
\end{table}

\begin{table}[!htb]
    \renewcommand{\arraystretch}{1.2} 
    \centering
    \caption{Optimal clustering results for each value of min\_samples, with BCubed metrics computed using the intent labels.}
    \resizebox*{\textwidth}{!}{%
    \begin{tabular}{@{}lccccccccccccc@{}} 
    \hline
    & & & \phantom{abc} & \multicolumn{3}{c}{Clusters} & \phantom{abc} & \multicolumn{3}{c}{Soft Clusters} & \phantom{abc}& \\
    \cline{5-7}
    \cline{9-11}
    {min\_samples} & min\_cluster\_size & validity && P & R & M && P & R & M && n clusters\\
    \hline
    \textbf{10} & \textbf{50} & \num{4.31e-2} && 0.1739 &	0.6529 & \textbf{0.2746} && 0.3231 & 0.1633 & 0.2170 &&	\textbf{105}\\
    20 & 25 & \num{4.01e-2} && 0.1521 & 0.7269 & 0.2516 && 0.3324 & 0.1574 & 0.2136 && 159\\
    30 & 250 & \num{4.38e-2} && 0.1314 & 0.6662 & 0.2196 && 0.2008 & 0.4272 & 0.2732 && 16\\
    \textbf{40} & \textbf{225} & \num{4.23e-2} && 0.1294 & 0.6825 & 0.2176 && 0.1960 & 0.4609 & \textbf{0.2751} && \textbf{15}\\
    50 & 50 & \num{2.48e-2} && 0.1242 & 0.7437 & 0.2128 && 0.2655 & 0.2634 & 0.2644 && 55\\
    60 & 50 & \num{2.06e-2} && 0.1191 & 0.7339 & 0.2050 &&	0.2345 & 0.3192 & 0.2704 && 43\\
    70 & 300 & \num{3.09e-2} && 0.1130 & 0.7161 & 0.1953 && 0.1819 & 0.4402 & 0.2574 && 11\\
    80 & 300 & \num{1.04e-1} && 0.1049 & 0.7034 & 0.1825 && 0.1678 & 0.4697 & 0.2473 && 10\\
    90 & 250 & \num{9.97e-2} && 0.1090 & 0.6937 & 0.1885 && 0.1756 & 0.4608 & 0.2543 && 12\\
    \textbf{100} & \textbf{300} & \num{1.50e-1} && 0.1043 & 0.6910 & 0.1812 && 0.1605 & 0.4755 & 0.2400 && \textbf{9}\\
    \hline
    \end{tabular}
    }%
   \label{tab:finetune_allintents_mwoz}
\end{table}

For a deeper analysis of the possible clustering results, we present the values for BCubed precision (P), BCubed recall (R) and their harmonic mean (M), for each value of \verb|min_samples| and the corresponding optimal value of \verb|min_cluster_size|. the values for BCubed metrics using domains or intents as labels are presented in Tables \ref{tab:finetune_alldomains_mwoz} and \ref{tab:finetune_allintents_mwoz}, respectively. Besides, we can evaluate both the clusters and soft clusters, where the latter are obtained by choosing the cluster with the maximum value of probability. The number of obtained clusters for each experience is also presented, allowing to have an idea of how granular the clusters are.

We can draw a few ideas from the results. Firstly, that an increase in the value of P is usually combined with a decrease in the value of R, supporting the need for analysing their harmonic mean (M). We can also confirm that we need to increase both \verb|min_samples| and \verb|min_cluster_size| for the clustering to become more conservative: for the same value of \verb|min_cluster_size|, an increase in \verb|min_samples| leads to a lower number of obtained clusters (which happens for $\verb|min_cluster_size|=50$, for example). 

\begin{figure}[!htb]
	\subfloat[Clusters for min\_samples=100 and min\_cluster\_size=300.
	\label{subfig:cluster_alldomains_multiwoz}]{\includegraphics[width=0.49\linewidth]{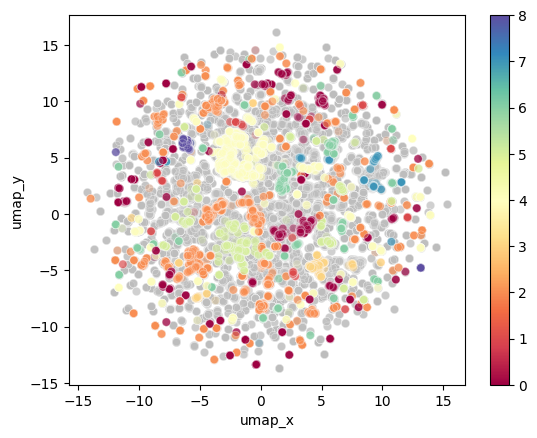}}%
	\subfloat[Soft clusters for min\_samples=100 and min\_cluster\_size=300.
	\label{subfig:soft_cluster_alldomains_multiwoz}]{\includegraphics[width=0.49\linewidth]{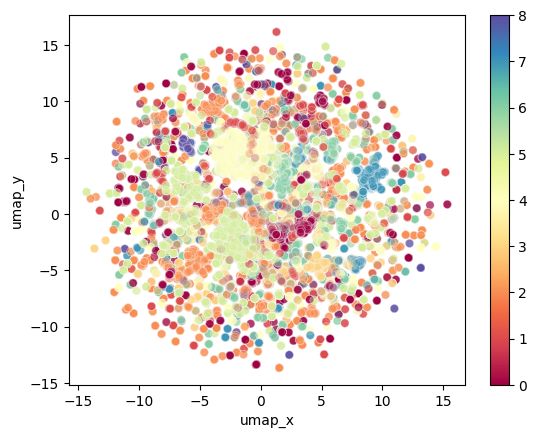}}%
	\caption{Results of clustering domains in MultiWOZ {in MultiWOZ using optimal samples and cluster size by intrinsic measures}.}
	\label{fig:clustering_alldomains_multiwoz}
\end{figure}

The BCubed metrics results are generally better when using the domain annotations as labels. In Figure \ref{fig:clustering_alldomains_multiwoz}, we present the results for the inter-domain experience with the optimal relative validity index, where the quality of the clusters can be grasped. In Table \ref{tab:alldomains_mwoz_frequentwords}, we present details about each cluster: their length, persistence in the spanning tree, and the dataset reference label, which corresponds to the label from the dataset with more data points in each cluster. For a better analysis of each clustering experience, we can also extract the most frequent words in each cluster of utterances. In this experience, we use the TD-IDF algorithm, treating each cluster of utterances as a single document\footnote{Following \url{https://towardsdatascience.com/topic-modeling-with-bert-779f7db187e6}}.

\begin{table}[!htb]
    \renewcommand{\arraystretch}{1.2} 
    \centering
    \caption{Details of the clusters obtained for all the domains.}
    \resizebox*{0.9\textwidth}{!}{%
    \begin{tabular}{lcccc}
    \hline
    cluster & length & persistence & top words {by TF-IDF} & label \\
    \hline
        0 & 300 & 0.0829 & postcode, phone, address, number, code & attraction\\
        1 & 300 & 0.0197 & price, range, preference, options, area & hotel\\
        2 & 674 & 0.0578 & train, time, cambridge, leave, leaving &  train\\
        3 & 451 & 0.0464 & taxi, need, time, cardinal, contact & taxi\\
        4 & 321 & 0.0586 & guesthouse, hotel, free, parking, star & hotel\\
        5 & 300 & 0.0365 & restaurant, food, centre, town, restaurants & restaurant\\
        6 & 314 & 0.0445 & people, date, cardinal, book, yes & hotel\\
        7 & 300 & 0.0549 & reference, number, booking, successful, booked & booking general\\
        8 & 300 & 0.1402 & fee, gbp, total, station, payable & general train\\
    \hline
    \end{tabular}
    }%
   \label{tab:alldomains_mwoz_frequentwords}
\end{table}

It is possible to say that the algorithm is successfully identifying different clusters of domains, as most of the obtained clusters are clearly from the domain assigned as label. While a few others seem to be more general (clusters 0, 1 and 6), we understand that these types of utterances must have a great presence in the dataset, and possibly appearing in different types of domain dialogues. We should underline that, as the amount and variability of dialogue utterances increase, it is more likely that similar utterances belonging to different domains appear, leading to utterances with different labels being clustered together.

\subsection{Intra-domain Clustering}
\label{subsec:intradomain_clustering}

For this experience, we consider utterances from the MultiWOZ dataset belonging to the hotel domain. Intra-domain data is the most likely to be found in a real world scenario, where dialogues that are jointly analyzed belong to the same broader domain.

In Table \ref{tab:finetune_hotel_validity}, the values for the relative validity index are presented when varying \verb|min_samples| from 5 to 50 with a step size of 5, and \verb|min_cluster_size| from 10 to 100 with a step size of 10 --- as we are in the presence of a smaller amount of data, the range of values for the variables have also been decreased. The best score of relative validity index is for the combination of $\verb|min_samples|=50$ and $\verb|min_cluster_size|=80$. 

\begin{table}[!htb]
  \renewcommand{\arraystretch}{1.2} 
  \centering
  \caption{Grid search over min\_cluster\_size and min\_samples for the hotel domain.}
  \resizebox*{\textwidth}{!}{%
  \begin{tabular}{lccccccccccc}
    \hline
     & 10 & 20 & 30 & 40 & 50 & 60 & 70 & 80 & 90 & 100 \\
    \hline
        5 & \num{5.91e-2} &	\num{5.45e-2} & \num{5.77e-5} & \num{2.89e-5} & \num{2.89e-5} & \num{1.35e-5} & \num{1.35e-5} & \num{1.35e-5} & \num{1.35e-5} & \num{1.35e-5} \\
        10 & \num{4.29e-2} & \num{5.36e-2} & \num{1.23e-3} & \num{2.92e-5} & \num{2.96e-5} & \num{2.96e-5} & \num{2.96e-5} & \num{2.96e-5} & \num{2.96e-5} & \num{5.88e-5} \\
        15 & \num{3.14e-2} & \num{2.78e-2} & \num{2.86e-2} & \num{2.27e-2} & \num{2.25e-2} & \num{4.92e-5} & \num{3.21e-5} & \num{3.21e-5} & \num{3.21e-5} & \num{3.21e-5} \\
        20 & \num{3.36e-2} & \num{2.69e-2} & \num{1.02e-5} & \num{1.02e-5} & \num{1.02e-5} & \num{4.21e-3} & \num{4.21e-3} & \num{1.41e-6} & \num{1.41e-6} & \num{1.41e-6}\\
        25 & \num{2.69e-2} & \num{2.99e-2} & \num{6.98e-7} & \num{6.98e-7} & \num{6.66e-3} & \num{6.66e-3} & \num{6.66e-3} & \num{5.54e-8} & \num{5.54e-8} &	\num{5.54e-8}\\
        30 & \num{5.41e-4} & \num{3.52e-6} & \num{3.92e-6} & \num{3.92e-6} & \num{1.35e-2} & \num{1.35e-2} & \num{1.35e-2} & \num{1.09e-2} & \num{7.53e-3} & \num{1.79e-5}\\
        35 & \num{5.97e-4} & \num{1.89e-6} & \num{6.47e-3} & \num{6.47e-3} & \num{6.47e-3} & \num{6.47e-3} & \num{6.47e-3} &	\num{1.86e-7} & \num{1.86e-7} & \num{1.86e-7}\\
        40 & \num{8.81e-4} & \num{2.95e-5} & \num{5.67e-3} & \num{5.67e-3} & \num{5.67e-3} & \num{5.67e-3} & \num{5.67e-3} & \num{5.67e-3} &	\num{2.92e-6} & \num{2.92e-6}\\
        45 & \num{6.33e-4} & \num{7.52e-3} & \num{7.52e-3} & \num{7.52e-3} & \num{7.52e-3} & \num{7.53e-3} & \num{5.17e-3} & \num{2.70e-3} & \num{4.46e-6} & \num{3.48e-6}\\
        50 & \num{2.09e-5} & \num{2.09e-5} & \num{2.09e-5} & \num{2.09e-5} & \num{2.09e-5} & \num{2.09e-5} & \num{1.10e-2} & \num{3.11e-1} & \num{3.11e-1} & \num{4.71e-6}\\

    \hline
  \end{tabular}
  }%
  \label{tab:finetune_hotel_validity}
\end{table}

\begin{table}[!htb]
    \renewcommand{\arraystretch}{1.2} 
    \centering
    \caption{Optimal clustering results for each value of min\_samples.}
    \resizebox*{\textwidth}{!}{%
    \begin{tabular}{@{}lccccccccccccc@{}} 
    \hline
    & & & \phantom{abc} & \multicolumn{3}{c}{Clusters} & \phantom{abc} & \multicolumn{3}{c}{Soft Clusters} & \phantom{abc}& \\
    \cline{5-7}
    \cline{9-11}
& min\_cluster\_size & validity && P & R & M && P & R & M && n clusters\\
    \hline
    5 & 10 & \num{5.91e-2} && 0.5509 &	0.6357 & 0.5903 &&	0.6527 & 0.0381 & 0.0721 && 161\\
    10 & 20 & \num{5.36e-2} && 0.5282 & 0.7183 & 0.6088 && 0.6164 & 0.0703 & 0.1262 &&	59\\
    15 & 10 & \num{3.14e-2} && 0.5200 & 0.7790 & 0.6237 && 0.6344 & 0.0525 & 0.0970 &&	96\\
    20 & 10 & \num{2.69e-2} && 0.5155 &	0.7694 & 0.6173 &&	0.6038 & 0.0870 & 0.1520 &&	50\\
    \textbf{25} & \textbf{20} & \num{2.99e-2} && 0.5158 &	0.8127 & \textbf{0.6311} && 0.5725 & 0.1055 & 0.1781 && \textbf{28}\\
    30 & 50 & \num{1.35e-2} && 0.4994 & 0.5578 & 0.5270 && 0.5129 & 0.7407 & 0.6061 &&	6\\
    35 & 30 & \num{6.47e-3} && 0.4971 & 0.5491 & 0.5218 && 0.5117 & 0.7548 & 0.6100	&& 6\\
    40 & 30 & \num{5.67e-3} && 0.4956 &	0.5368 & 0.5154 &&	0.5125 & 0.7593 & 0.6120 &&	6\\
    \textbf{45} & \textbf{20} & \num{7.52e-3} && 0.4964 & 0.5272 & 0.5113 &&	0.5128 & 0.7743 & \textbf{0.6170} &&	\textbf{6}\\
    \textbf{50} & \textbf{80} & \textbf{\num{3.11e-1}} && 0.4885 & 0.5328 & 0.5097 &&	0.4921 & 0.8209	& 0.6153 && \textbf{3}\\
    \hline
    \end{tabular}
    }%
   \label{tab:finetune_hotel}
\end{table}

\begin{figure}[htb!]
	\centering
	\subfloat[Clusters for min\_samples=50 and min\_cluster\_size=80.
	\label{subfig:cluster_hotel_3clusters}]{\includegraphics[width=0.45\linewidth]{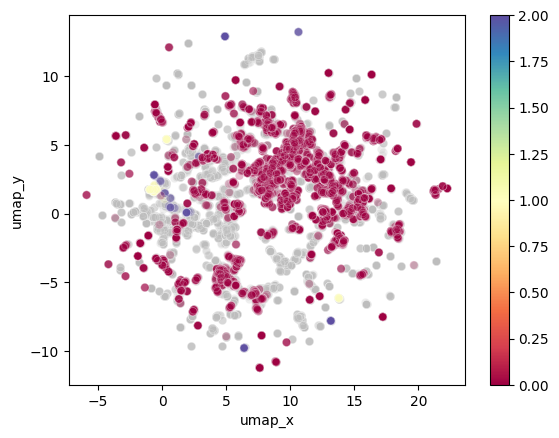}}%
	\subfloat[Soft clusters for min\_samples=50 and min\_cluster\_size=80.
	\label{subfig:soft_cluster_hotel_3clusters}]{\includegraphics[width=0.45\linewidth]{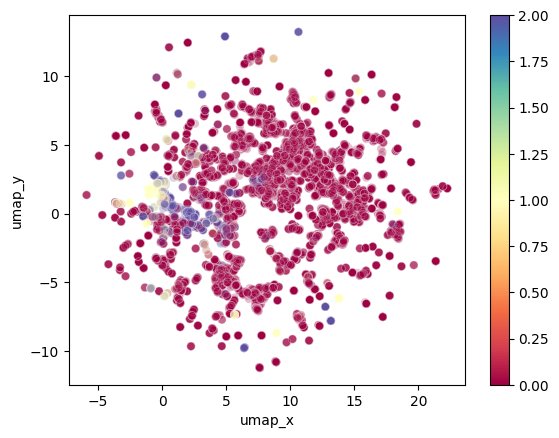}}%
	
	\subfloat[Clusters for min\_samples=45 and min\_cluster\_size=20.
	\label{subfig:cluster_hotel_6clusters}]{\includegraphics[width=0.45\linewidth]{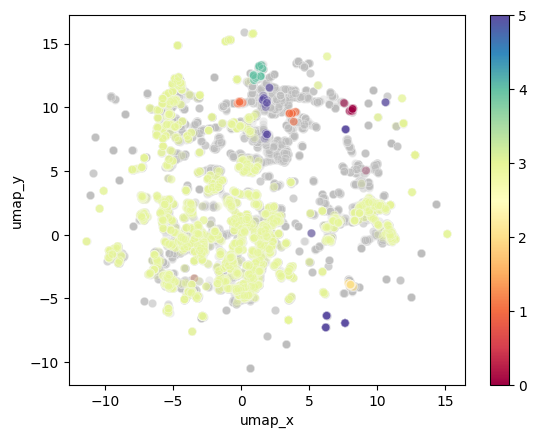}}%
	\subfloat[Soft clusters for min\_samples=45 and min\_cluster\_size=20.
	\label{subfig:soft_cluster_hotel_6clusters}]{\includegraphics[width=0.45\linewidth]{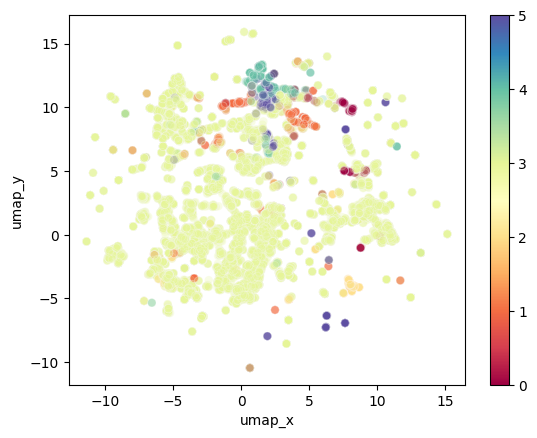}}%
	
	\subfloat[Clusters for min\_samples=25 and min\_cluster\_size=20.
	\label{subfig:cluster_hotel_28clusters}]{\includegraphics[width=0.45\linewidth]{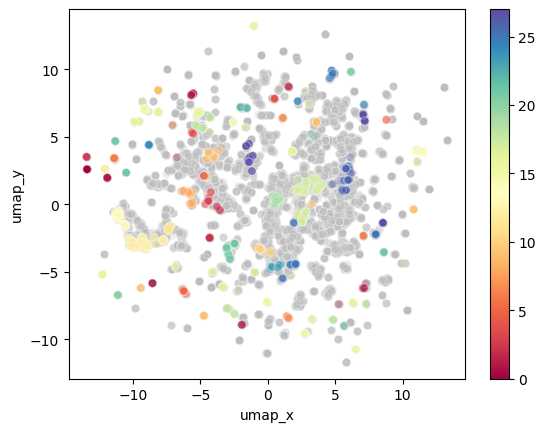}}%
	\subfloat[Soft clusters for min\_samples=25 and min\_cluster\_size=20.
	\label{subfig:soft_cluster_hotel_28clusters}]{\includegraphics[width=0.45\linewidth]{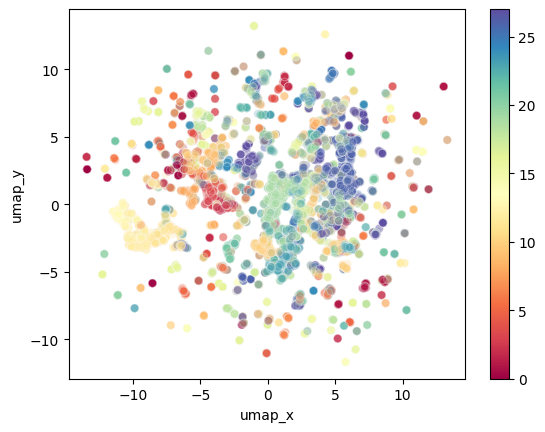}}%
	\caption{Results of clustering intents in the hotel domain.}
	\label{fig:clustering_hotel_domain}
\end{figure}

Similarly to before, we present the values P, R, and M in Table \ref{tab:finetune_hotel}, for each value of \verb|min_samples| and the corresponding optimal value of \verb|min_cluster_size|. In what comes to the best performance in BCubed metrics, there is a mismatch between the results from the clusters and soft clusters: the former occurs for $\verb|min_samples|=25$ and $\verb|min_cluster_size|=20$; and the latter is for $\verb|min_samples|=45$ and $\verb|min_cluster_size|=20$. These results are also not in accordance with the optimal performance for relative validity index, which happens for $\verb|min_samples|=50$ and $\verb|min_cluster_size|=80$. From these possible combinations of values, 28 is the amount of obtained clusters which is more in accordance with the original labels from the dataset.

In Figure \ref{fig:clustering_hotel_domain}, we present the results of these three clustering experiences in a 2-D representation, from fewer to greater obtained clusters. {The color gradient on the right side of each graph indicates the number of clusters present in the plot, where the top indicates the maximum number of clusters +1.} For experiences where fewer clusters are obtained, there is generally a broader cluster to which most of the data points belong, with a few more specific ones. Although this can be supported by the nature of the dialogues, where a lot of utterances are related to searching for a hotel, these results are not that useful once we want to analyse the flow of intentions in a dialogue. This fact advocates for the importance of adapting the hyper parameters to the experience and results we are looking for, regardless of any computed metric. In Tables \ref{tab:hotel_3clusters_frequentwords},  \ref{tab:hotel_6clusters_frequentwords} and  \ref{tab:hotel_28clusters_frequentwords}, we present details about each cluster, for each clustering experience of Figure \ref{fig:clustering_hotel_domain}.

\begin{table}[!htb]
    \renewcommand{\arraystretch}{1.2} 
    \centering
    \caption{Details of the 3 clusters obtained for the hotel domain.}
    \resizebox*{0.8\textwidth}{!}{%
    \begin{tabular}{lcccc}
    \hline
    cluster & length & persistence & top words & label \\
    \hline
        0 & 471 & 0.0526 & hotel, guesthouse, date, cardinal, free & hotel-inform\\
        1 & 80 & 0.0213 & range, price, moderate, cheap, don & hotel-inform\\
        2 & 82 & 0.0112 & price, range, options, cardinal, preference & hotel-request\\
    \hline
    \end{tabular}
    }%
   \label{tab:hotel_3clusters_frequentwords}
\end{table}

\begin{table}[!htb]
    \renewcommand{\arraystretch}{1.2} 
    \centering
    \caption{Details of the 6 clusters obtained for the hotel domain.}
    \resizebox*{0.8\textwidth}{!}{%
    \begin{tabular}{lcccc}
    \hline
    cluster & length & persistence & top words {by TF-IDF} & label \\
    \hline
        0 & 20 & 0.0674 & reference, number, yes, need, book & hotel-request\\
        1 & 20 & 0.0181 & postcode, phone, address, number, code & hotel-request\\
        2 & 20 & 0.0848 & restaurant, taxi, hotel, time, need & hotel-inform\\
        3 & 309 & 0.0397 & cardinal, guesthouse, hotel, date, free & hotel-inform\\
        4 & 20 & 0.0421 & range, price, moderate, cheap, priced & hotel-inform\\
        5 & 24 & 0.0465 & price, range, options, mind, area & hotel-request\\
    \hline
    \end{tabular}
    }%
   \label{tab:hotel_6clusters_frequentwords}
\end{table}

\begin{table}[!htb]
    \renewcommand{\arraystretch}{1.2} 
    \centering
    \caption{Details of the 28 clusters obtained for the hotel domain.}
    \resizebox*{\textwidth}{!}{%
    \begin{tabular}{lcccc}
    \hline
    cluster & length & persistence & top words {by TF-IDF} & label \\
    \hline
        0 & 20 & \num{2.00e-9} & yes, does, fine, sounds, matter & hotel-inform\\
        1 & 20 & \num{1.06e-7} & date, time, try, starting, instead & hotel-inform\\
        2 & 21 & \num{3.39e-8} & phone, number, postcode, date, help & hotel-inform\\
        3 & 20 & \num{1.01e-7} & postcode, phone, number, just, address & hotel-request\\ 
        4 & 20 & \num{3.28e-8} & address, road, phone, number, town & hotel-request\\
        5 & 21 & \num{3.81e-7} & restaurant, taxi, hotel, time, need & hotel-inform\\
        6 & 21 & \num{1.48e-7} & book, reference, number, yes, sounds & hotel-request\\
        7 & 21 & \num{2.39e-7} & reference, number, yes, need, thank & hotel-request\\
        8 & 20 & \num{1.57e-7} & range, price, moderate, cheap, priced & hotel-inform\\
        9 & 20 & \num{1.70e-7} & price, range, options, mind, area & hotel-request\\
        10 & 20 & \num{2.47e-8} & hotels, hotel, sorry, area, criteria & hotel-nooffer hotel-request\\
        11 & 20 & \num{2.85e-8} & date, people, starting, room, cardinal & hotel-inform\\
        12 & 46 & \num{2.28e-7} & date, people, starting, book, cardinal & hotel-inform\\
        13 & 20 & \num{1.97e-7} & date, people, starting, cardinal, yes & hotel-inform\\
        14 & 26 & \num{3.99e-1} & wifi, does, free, internet, include & hotel-inform\\
        15 & 22 & \num{8.29e-7} & parking, free, does, offer, yes & hotel-inform\\
        16 & 21 & \num{6.88e-7} & area, stay, town, like, prefer & hotel-request\\
        17 & 20 & \num{8.89e-8} & hotel, prefer, preference, guesthouse, hotels & hotel-inform hotel-request\\ 
        18 & 22 & \num{5.30e-7} & place, stay, looking, need, north & hotel-inform\\
        19 & 20 & \num{8.83e-8} & guesthouse, cardinal, star, like, stars & hotel-inform\\ 
        20 & 33 & \num{7.69e-7} & guesthouse, lovely, does, tell, house & hotel-recommend\\
        21 & 22 & \num{6.40e-7} & called, hotel, looking, guesthouse, information & hotel-inform\\
        22 & 20 & \num{2.87e-7} & guesthouse, suggest, recommend, prefer, like & hotel-recommend\\
        23 & 20 & \num{5.23e-8} & guesthouse, book, like, room, recommend & hotel-recommend\\
        24 & 21 & \num{3.90e-9} & parking, place, stay, free, cheap & hotel-inform\\
        25 & 20 & \num{3.30e-7} & parking, guesthouse, free, looking, cheap & hotel-inform\\
        26 & 21 & \num{1.60e-7} & star, cardinal, hotel, free, rating & hotel-inform\\
        27 & 40 & \num{1.07e-7} &  wifi, free, parking, need, hotel & hotel-inform\\
    \hline
    \end{tabular}
    }%
   \label{tab:hotel_28clusters_frequentwords}
\end{table}

For the experience with only 3 obtained clusters (Table \ref{tab:hotel_3clusters_frequentwords}), it is easy to understand that the two specific clusters are related to the hotel prince range: cluster 1 (yellow) is probably mostly composed of utterances from the user, due to the high presence of restrictive words (`moderate' and `cheap'); cluster 2 (purple) should be mostly composed of utterances from the assistant where a `preference' is recurrently being asked. The rest of the utterances belong to cluster 0 (magenta), where the most frequent words are certainly directly obtained from the most frequent utterances from the dataset. 

In the next experience (Table \ref{tab:hotel_6clusters_frequentwords}), there are other more specific clusters, regarding booking (cluster 0 - magenta), hotel details such as postcode, phone, and address (cluster 1 - orange), and requesting a taxi from the hotel to the restaurant (cluster 2 - dark yellow). 

The last experience results in a higher number of clusters, spanning over more versatile types of intents: a confirmation (cluster 0), a suggestion of other time or date (cluster 1), a recognition of the non existence of hotels following the given criteria (cluster 10), an inquiry about the wifi (cluster 14), etc. The fact that the clusters are more granular also means that the algorithm can split some clusters that could be broader, such as cluster 11 and 12, which both seem to be about a hotel room booking request. One possibility can be the fact that one cluster includes more utterances belonging to user inquiries, and the other to assistant replies.

In the three clustering experiences, most of the clusters are labelled with either `hotel-inform' or `hotel-request', which are the most frequent labels of utterances in the hotel domain, as seen in Figure \ref{fig:hotel_intents_embeddings}. We can understand that, despite being able to obtain reasonable clusters, it will be difficult for the algorithm to match the level of granularity with the dataset annotations, which explains the low results for the BCubed metrics.

\subsection{Analysis of the dialogue flow}

For this part of the experience, we feed the results from the intra-domain clustering of the hotel domain to the tool for analysis of sequences. In Table \ref{tab:hotel_sequence_analysis}, the most frequent flows {between these 28 clusters} are presented, which can be informally analysed resorting to the most relevant utterances in each cluster. 

\begin{table}[!htb]
    \renewcommand{\arraystretch}{1.2} 
    \centering
    \caption{The most frequent sequences of {the identified 28} clusters for the hotel domain.}
    \resizebox*{0.3\textwidth}{!}{%
    \begin{tabular}{ccc}
    \hline
    n & sequence & frequency \\
    \hline
        2 & $26 \rightarrow 19$ & 767\\
        2 & $19 \rightarrow 12$ & 625\\
        2 & $10 \rightarrow 19$ & 621\\
        2 & $26 \rightarrow 10$ & 574\\
        2 & $27 \rightarrow 19$ & 559\\
        2 & $26 \rightarrow 12$ & 492\\
        2 & $26 \rightarrow 23$ & 492\\
        2 & $19 \rightarrow 11$ & 451\\
        2 & $19 \rightarrow 23$ & 435\\
        2 & $21 \rightarrow 19$ & 420\\
        3 & $26 \rightarrow 10 \rightarrow 19$ & 249 \\
        3 & $26 \rightarrow 19 \rightarrow 12$ & 204 \\
        3 & $10 \rightarrow 19 \rightarrow 12$ & 166 \\
        3 & $27 \rightarrow 10 \rightarrow 19$ & 162 \\
        3 & $27 \rightarrow 19 \rightarrow 12$ & 161 \\
        3 & $26 \rightarrow 10 \rightarrow 12$ & 156 \\
        3 & $27 \rightarrow 26 \rightarrow 19$ & 155 \\
        3 & $10 \rightarrow 26 \rightarrow 19$ & 151 \\
        3 & $26 \rightarrow 10 \rightarrow 23$ & 141 \\
        3 & $26 \rightarrow 19 \rightarrow 23$ & 141
        \\
    \hline
    \end{tabular}
   \label{tab:hotel_sequence_analysis}
    }
\end{table}

\begin{figure}[!htb]
  \centering
  \includegraphics[width=0.8\textwidth]{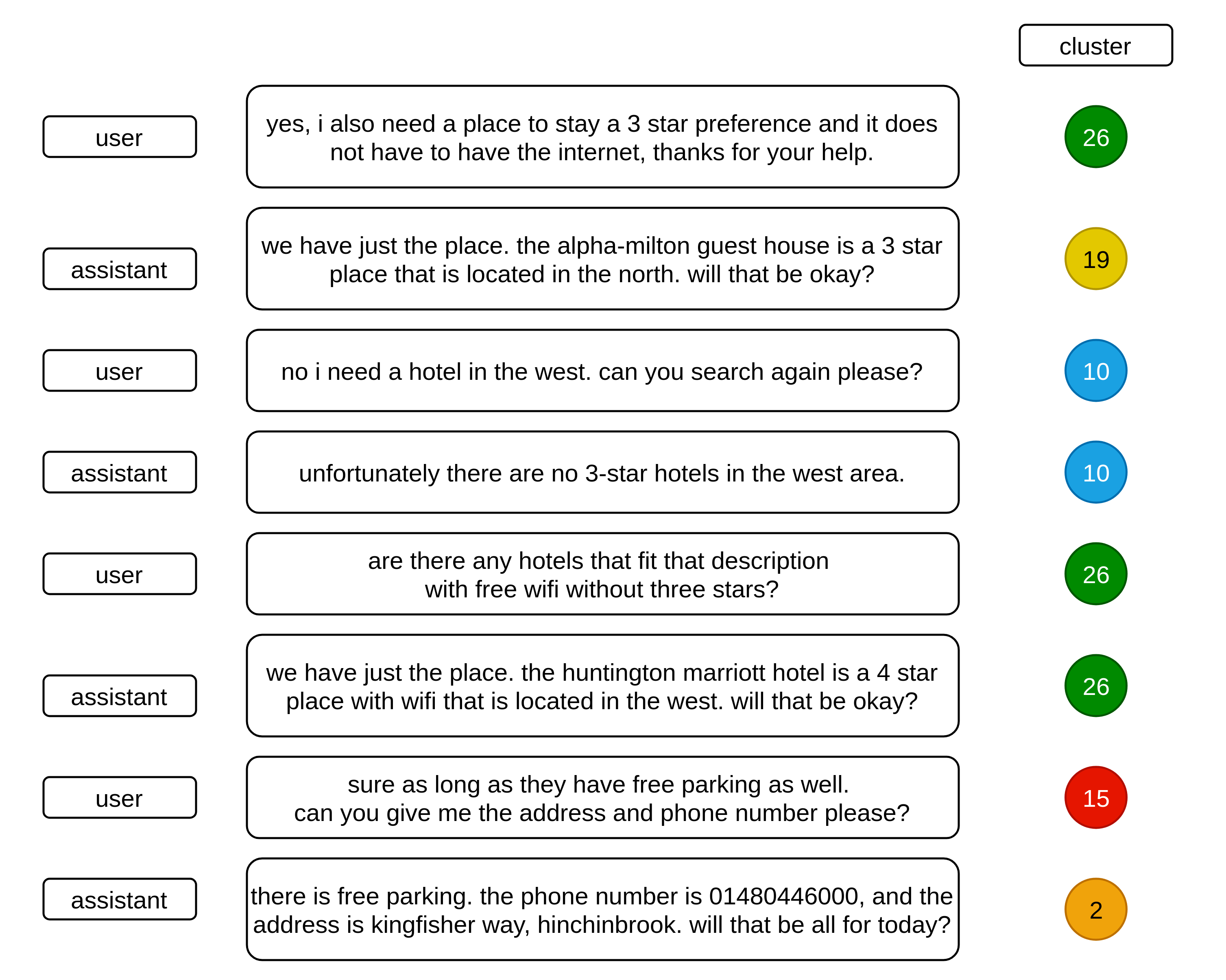}
  \caption{A dialogue example with the assigned clusters.}
  \label{fig:dialogue_clusters_example}
\end{figure}

Clusters 26 and 27 appear frequently, which are composed of utterances where the user is asking for a hotel with some specific restrictions: the former with the intent for a particular star rating, and the latter with parking or/and wifi restrictions. Afterwards, the most common clusters are 10 and 19: cluster 10 identifies the lack of domain entities obeying to the given specifications; and cluster 19 suggests a hotel or guesthouse. Cluster 12 is also frequent, usually assigned to utterances where the user is starting the booking process.

Despite being possible to make this correspondence, some cases do not follow these labels, such as the transition $10 \rightarrow 19$, that apparently matches two subsequent assistant utterances. As the utterances from the user and assistant are all clustered at the same time, semantically similar utterances from both of the parts can be assigned the same cluster. However, this experience was not focused on dividing the utterances between user and system, as this also does not happen in the dataset reference labels: as an example, there are a lot of `hotel-inform' subsequent utterances.

As an example, we provide a dialogue example with the assigned clusters, in Figure \ref{fig:dialogue_clusters_example}. The dialogue starts with the transition $26 \rightarrow 19$, which is the most common transition in the dataset. Afterwards, it classifies two subsequent utterances with the cluster 10, which can be justified by being semantically close (both present negative sentences). The user comes back to providing hotel restrictions, which is aligned with what we have seen about cluster 26. The following suggestion from the assistant (the $6^{th}$ utterance) is also assigned to the cluster 26, which is not aligned with what we discovered about the clusters --- it should probably be assigned with cluster 19. One justification for these errors can be, that as we are forcing the algorithm to assign one cluster to each utterance {(as we used the results from soft-clustering)}, very weak classifications are also being considered. Besides, the most frequent clusters should also be the ones that are not that specific, and the algorithm has more difficulties in classifying. When it comes to the booking itself, the algorithm assigns two different clusters for asking and providing the requirements, 15 and 2, which are in accordance with the main topics extracted from the clusters: the first one is confirming the hotel has free parking, and the latter providing the required hotel details.

\section{Conclusion and Future Work}\label{section:ConclusionsAndFutureWork}

In this work, we successfully built a framework that is able to identify dialogue intentions, in an unsupervised manner. To do so, we developed a clustering tool for dialogue utterances, which groups them according to their similarity and intention. As seen in the experiments, we were able to obtain reasonable clusters with different levels of granularity, supporting the idea that the algorithm parameters should be adapted to each use case and nature of the data, regardless of how general the algorithm should be.

Besides, the sequence analysis tool proved to be able to find relevant flows of intentions in a dialogue, which can be helpful for dialogue management applications. In future work, it would make sense to perform two different clustering experiences, for user and assistant utterances apart, to ensure they are not mixed in the same clusters. {Depending on the application, this information could even be available, and an analysis of the sequence of user requests without the assistant (and vice versa) could be valuable.}

Besides, the problem of identifying dialogue flows can be further investigated by modifying the sequence analysis tool to return sequences obeying to different specifications, such as a longer length or sequences that do not include a certain cluster. Regardless of that, these results do already prove that it is possible to identify relevant clusters in a dialogue application, and analyse their most common flows in an unsupervised scenario. Other opportunities of future work are the creation of a taxonomy of intents, and the comparison with the one provided in the datasets.

\section*{Acknowledgements}

This work was conducted within the IAG (Intelligent
Agents Generator) project with the universal code
LISBOA-01-0247-FEDER-045385, co-funded by Lisboa 2020, Portugal 2020, and the European Union, through the European Regional Development Fund.

\bibliographystyle{nlelike}
\bibliography{library}

\label{lastpage}

\end{document}